\documentclass[nonblindrev]{informs3} 

\usepackage{balance}
\usepackage{amsmath}
\usepackage[tight,footnotesize]{subfigure}
\usepackage{graphicx}
\usepackage{bm}
\usepackage{algorithm}
\usepackage[noend]{algorithmic}
\usepackage{multirow}
\usepackage{slashbox}
\usepackage{caption}
\usepackage{amsfonts}

\OneAndAHalfSpacedXII 

\usepackage{natbib}
 \bibpunct[, ]{(}{)}{,}{a}{}{,}%
 \def\bibfont{\small}%
\TheoremsNumberedThrough     

\EquationsNumberedThrough    

\begin{document}


\RUNAUTHOR{Tang et al.}

\RUNTITLE{Adaptive Submodular Meta-Learning}

\TITLE{Adaptive Submodular Meta-Learning}

\ARTICLEAUTHORS{%
\AUTHOR{Shaojie Tang}
\AFF{Naveen Jindal School of Management, The University of Texas at Dallas}
\AUTHOR{Jing Yuan}
\AFF{Department of Computer Science, The University of Texas at Dallas}
} 

\ABSTRACT{Meta-Learning has gained increasing attention in the machine learning and artificial intelligence communities.  In this paper, we introduce and study an adaptive submodular meta-learning problem. The input of our problem is a set of items, where each item has a random state which is initially unknown. The only way to observe an item's state is to select that item. Our objective is to adaptively select a group of items that achieve the best performance over a set of tasks, where each task is represented as an adaptive submodular function that maps sets of items and their states to a real number. To reduce the computational cost while maintaining a personalized solution for each future task, we first select an initial solution set based on previously observed tasks, then adaptively add the remaining items to the initial solution set when a new task arrives. As compared to the solution where a brand new solution is computed for each new task, our meta-learning based approach leads to lower computational overhead at test time since the initial solution set is pre-computed in the training stage.  To solve this problem, we propose a two-phase  greedy policy and show that it achieves a $1/2$ approximation ratio for the monotone case. For the non-monotone case, we develop a two-phase randomized greedy policy that achieves a $1/32$ approximation ratio.}


\maketitle

\section{Introduction}
The goal of meta-learning is to leverage a few training examples to improve the performance of the learning algorithm on  future tasks \citep{thrun2012learning}. Among numerous  formulations for meta-learning,  Model-Agnostic Meta-Learning \citep{finn2017model} is one of the most popular ones in continuous domain. MAML aims to provide a good initialization of a model's parameters that can be quickly adapted to a new task using only a small number of gradient steps. \cite{adibi2020submodular} extend the methodology of MAML to the discrete domain and introduce the submodular meta-learning problem. Under their setting, each task is presented as a monotone and submodular utility function and their goal is to select a group of $k$ items that achieves the best performance over all tasks. Their submodular meta-learning framework can be done in two parts: They first select an initial set of items based on some observed tasks, then after observing a new task, they  add some additional items to that initial set to build a personalized solution for each new task. Their approach can find a personalized solution for each new task while reducing the computational overhead at test time. This is because the first part, which finds a good initial solution set, is done offline, it does not consume any resource at test time.

In this paper, we extend their study to the adaptive setting. Under our setting, each item has a random state drawn from some known prior distribution. Initially, each item's state is unknown, we must select an item in order to observe its realized state. Our goal is to adaptively select a group of $k$ items for each incoming task so as to maximize the average expected utility over all tasks. We assume that each task can be represented as an adaptive submodular function that maps sets of items and their states to a real number.  Consider the example of adaptive viral marketing \citep{golovin2011adaptive}, where we would like to promote a product through a social network.  Suppose that we have data on a social network where nodes represent individuals and edges represent social relations,  our objective is to choose influential sets of individuals. In this context, items refer to individuals, the state of each item refers to the actual set of individuals it influences, and the promotion of a particular product can be considered as a task. As each product may have its own diffusion model that governs the diffusion process of this product,  it is reasonable to select different  influential sets of individuals for marketing different products. Hence, our objective is to adaptively select some individuals to trigger a large cascade of influence over all products. 

Following the framework of meta-learning, our adaptive submodular meta-learning is composed of two stages. In the first stage, we select an initial set of $l$ items non-adaptively based on prior experience. In the second stage, we adaptively add a group of additional $k-l$ items to the initial set after observing the incoming task. Our framework is general enough to capture numerous applications such as machine learning \citep{dasgupta2008hierarchical}, interactive recommendations \citep{karbasi2012comparison}, viral marketing \citep{yuan2017adaptive,yuan2017no}, and link prediction \citep{mitrovic2019adaptive}. Note that the two extremes of adaptive submodular meta-learning are (1) non-adaptive setting when $l=k$ (i.e., all $k$ selections are made before observing the incoming task),  and (2) fully adaptive setting when $l=0$ (i.e., all $k$ items are selected in a closed-loop manner after observing the incoming task where
each item is selected based on the feedback from previous selections). Clearly, there is trade off between the degree of personalization of our solution and the computational overhead at the test time. In particular, as $l$ decreases, we provide a more personalized solution for each incoming task, however, this also indicates that more selections need to be done adaptively at the test time, which may result in longer response time for a new task. Depending on the context of the application, the decision-maker can choose an appropriate $l$ that balances the computational overhead at test time  and the degree of personalization of the solution. We leave the selection of an appropriate $l$ to the decision-maker, while our focus is on finding the best learning policy for a fixed $l$. Our contributions are threefold:

$\bullet$ We develop a novel framework of adaptive submodular meta-learning where each item has a random state and each task can be represented using an adaptive submodular function. Our framework can find  numerous applications in machine learning and artificial intelligence.

$\bullet$ We show that the new objective function defined in our framework does \emph{not} satisfy the property of adaptive submodularity, despite the utility function of each task is adaptive submodular. This makes the existing results on adaptive submodular maximization not applicable to our setting. We overcome this challenge by proposing a series of effective policies for the adaptive  submodular meta-learning problem. Our policy enjoys the benefit of making selections adaptively while reducing the computational overhead at the test time. We show that our algorithm achieves a $1/2$ approximation ratio for the monotone case. For the non-monotone case, we develop a randomized policy that achieves a $1/32$ approximation ratio.

$\bullet$ We conduct extensive experiments to evaluate the performance of our solution for the application of adaptive viral marketing. Our results validate our theoretical analysis and show that the proposed solution outperforms all benchmark solutions.
\section{Related Work}
Meta-learning has been successfully applied to many domains, including reinforcement learning \citep{duan2016rl,fallah2020provably} and one-shot learning \citep{snell2017prototypical}. Model Agnostic Meta-learning \citep{finn2017model} is one of the most popular forms of meta-learning, it aims at learning an initial  model that can easily adapt to the new task from few examples. Most of existing studies, including MAML, consider the case where the feasible parameter space is continuous. Very recently, \cite{adibi2020submodular} extends this study to the discrete domain, i.e., they consider the case when the parameter space is discrete. Our study follows their work by considering a discrete variant of meta-learning. In \citep{adibi2020submodular}, they assume that each task can be represented using a monotone and submodular utility function. As a result, their objective is to find a good initial solution set that can quickly adapt to a new monotone and submodular function. In this work, we generalize their study by introducing an adaptive variant of submodular meta-learning. In particular, we assume that each item is associated with a random state whose realization is initially unknown. One must select an item in order to reveal its realized state.  The utility function of each task is defined over sets of items as well as their realized states. One natural approach to maximize a utility function under the above setting is to sequentially select a group of items, each selection is based on the feedback from previous selections. The previous submodular meta-learning framework falls short in adaptive settings as it requires the decision-maker to make selections regardless of the realization of items' states. To circumvent this issue, we adopt the notation of adaptive submodularity and adaptive monotone \citep{golovin2011adaptive}, which generalize the classic notations of submodularity and monotonicity from sets to policies. We assume that each task can be represented as an adaptive submodular function. Our study is also closely related to batch model active learning \citep{chen2013near}, where the selection is performed in batches. In each batch, they select multiple items in an offline fashion, and receive feedback only after all items from a batch have been selected. They develop a constant factor approximate solution to their problem. Their basic idea is to treat each possible batch as a virtual item, then apply the adaptive greedy algorithm over virtual items to obtain an approximate solution. Our work is different from theirs in two ways: As discussed in Section \ref{sec:111}, our utility function is not adaptive submodular, thus the standard analysis used in adaptive submodular maximization does not apply to our framework. Moreover, \cite{chen2013near} assume that each batch has the same size, while under our setting, we select $l$ items in the first batch (phase) and  each of the following batches contains a single item.
\section{Preliminaries}
We start by introducing some important notations. In the rest of this paper, we use  $[m]$ to denote the set $\{1, 2, \cdots, m\}$, and we use $|X|$ to denote the cardinality of a set $X$.

\subsection{Items and  States} We consider a set  $E$ of $n$ items. Each item is in a particular state from  $O$.  The item states are represented using a function $\phi: E\rightarrow O$, called a \emph{realization}. Hence, $\phi(e)$ represents the realization of $e$'s state. We use $\Phi=\{\Phi(e) \mid e\in E\}$ to represent the  random realizations of $E$, where $\Phi(e) \in O$ is a random realization of $e$. For any $Y\subseteq E$, let $\Phi(Y)= \cup_{e\in Y} \Phi(e)$ denote the random realizations of $Y$. There is a known prior probability distribution $p=\{\Pr[\Phi=\phi]: \phi\in U\}$ over realizations $U$.  The state $\Phi(e)$ of each item $e \in E$ is initially unknown, and we must select $e$ before observing  the value of $\Phi(e)$. After selecting a set of items, we are able to observe a \emph{partial realization} of those items' states.   For any partial realization $\psi$, we define the \emph{domain} $\mathrm{dom}(\psi)$ of $\psi$  the set of all items involved in $\psi$. We say a partial realization $\psi$ is consistent with a realization $\phi$, denoted $\phi \sim \psi$, if they are equal everywhere in $\mathrm{dom}(\psi)$. Moreover, we say $\psi$  is a \emph{subrealization} of  $\psi'$, denoted  $\psi \subseteq \psi'$, if $\mathrm{dom}(\psi) \subseteq \mathrm{dom}(\psi')$ and they are equal everywhere in the domain $\mathrm{dom}(\psi)$ of $\psi$. Let $p(\phi\mid \psi)$ denote the conditional distribution over realizations conditioned on  a partial realization $\psi$: $p(\phi\mid \psi) =\Pr[\Phi=\phi\mid \Phi\sim \psi ]$.

\subsection{Training and Test Tasks}  We consider a family $G$ of tasks, where the size of $G$ could be infinite. Each task $i\in G$ is represented via a utility function $f^i$ from a subset of items and their states to a non-negative real number: $f^i: 2^{E}\times O^E\rightarrow \mathbb{R}_{\geq 0}$.  We assume that tasks from $G$ arrive randomly  according to an underlying probability distribution $\theta$. Although  $\theta$  is not always available, we assume that we have observed a group  $M$ of $m$ training tasks. Each of the training tasks is sampled independently according to the distribution $\theta$. Our ultimate goal is to  maximize the utility function at test time. That is, we aim at achieving the best performance for an incoming task that is sampled independently from the distribution $\theta$.

Consider the task of adaptive viral marketing whose objective is to adaptively select a set of \emph{seed} users from a social network to help promote some product through the word-of-mouth effect. A social network can be represented as  a graph with vertices representing users and edges representing their relations. The information propagation process is governed by some product-specific stochastic cascade model. Notably, Independent Cascade model assigns a product-specific propagation probability to each edge, say $(u,v)$, and it represents the probability that user $u$ successfully promotes a product to user $v$. One approach for solving this problem is to find a fixed group of seed users that can generate the largest average influence over all observed products, and let them promote all other products arriving in the future. As this group of users is pre-computed offline, the aforementioned approach has zero computational overhead at test time, however, it fails to provide a personalized solution for each new task.  Another
approach is to adaptively select a group of seed users with respect to  each new task. Clearly, this approach leads to better performance, but at the cost of spending longer time and computational power on selecting all seed users at the test time.


\subsection{Policies} We consider an adaptive optimization problem where we sequentially select a group of items, after each selection, we observe the partial realization of the states of  those items which have been previously selected.  We define an adaptive policy using a function $\pi$ that maps a set of observations  to a distribution $\mathcal{P}(E)$ of $E$, specifying  which item to select next based on  the current task and the partial realization observed  so far: $\pi: 2^{E}\times O^E \times 2^M \rightarrow \mathcal{P}(E)$.

\begin{definition}[Policy  Concatenation]
Given two policies $\pi$ and $\pi'$,  let $\pi @\pi'$ denote a policy that runs $\pi$ first, and then runs $\pi'$, ignoring the observation obtained from running $\pi$.
\end{definition}

\begin{definition}[Level-$t$-Truncation of a Policy]
Given a policy $\pi$, we define its  level-$t$-truncation $\pi_t$  as a policy that runs $\pi$ until it selects $t$ items.
\end{definition}

 For each task $i\in G$ and each realization $\phi$, let $E(\pi, \phi, i)$ denote the subset of items selected by $\pi$ under realization $\phi$ for task $i$.  Note that $E(\pi, \phi, i)$ is a random variable. The expected  utility $f^i_{avg}(\pi)$ of a policy $\pi$ for task $i$ can be written as
$f^i_{avg}(\pi)=\mathbb{E}_{\Phi\sim p, \Pi}f^i(E(\pi, \Phi, i), \Phi)$, where the expectation is taken over the realization and the random output of the policy.  For any set of items $Y\subseteq E$ and any task $i\in G$, let $f^i(Y) = \mathbb{E}_{\Phi\sim p}f^i(Y, \Phi)$.



\subsection{Adaptive Submodularity and Monotonicity}
We next introduce several important notations.
\begin{definition}[Conditional Expected Marginal Utility of an Item]
\label{def:1}
Given a utility function $f^i$, the conditional expected marginal utility $\Delta_i(e \mid \psi)$ of an item $e$ conditioned on $\psi$ is
$\Delta_i(e \mid \psi)=\mathbb{E}_{\Phi}[f^i(\mathrm{dom}(\psi)\cup \{e\}, \Phi)-f^i(\mathrm{dom}(\psi), \Phi)\mid \Phi \sim \psi]$,
where the expectation is taken over $\Phi$ with respect to $p(\phi\mid \psi)=\Pr(\Phi=\phi \mid \Phi \sim \psi)$.
\end{definition}

\begin{definition}[Conditional Expected Marginal Utility of a Policy]
\label{def:1}
Given a utility function $f^i$, the conditional expected marginal utility $\Delta_i(\pi \mid \psi)$ of a policy $\pi$ conditioned on a partial realization $\psi$ is
$\Delta_i(\pi\mid \psi)=\mathbb{E}_{\Phi, \Pi}[f^i(\mathrm{dom}(\psi) \cup E(\pi, \Phi), \Phi)-f^i(\mathrm{dom}(\psi), \Phi)\mid \Phi\sim \psi]$, where the expectation is taken over (1) $\Phi$ with respect to $p(\phi\mid \psi)=\Pr(\Phi=\phi \mid \Phi \sim \psi)$, and (2) the random output of the policy. For any set of items $Y\subseteq E$, we define  the conditional expected marginal utility $\Delta_i(\pi \mid Y)$ of a policy $\pi$ conditioned on $Y$ as
\[\Delta_i(\pi\mid Y)=\mathbb{E}_{\Phi, \Pi}[f^i(Y \cup E(\pi, \Phi), \Phi)-f^i(Y, \Phi)\mid \Phi\sim p]\]
\end{definition}

We assume that the utility functions of all tasks in $M$ are \emph{adaptive submodular}  \citep{golovin2011adaptive}.  That is, for any two partial realizations $\psi$ and $\psi'$ such that $\psi\subseteq \psi'$, the following holds for each $i\in M$ and $e\in E\setminus \mathrm{dom}(\psi')$:
\begin{eqnarray}\label{def:22}
\Delta_i(e\mid \psi) \geq \Delta_i(e\mid \psi')
\end{eqnarray}
Moreover, we say a utility function $f^i$ is \emph{adaptive monotone} \citep{golovin2011adaptive} if  for any realization $\psi$, the following holds for each $e\in E\setminus \mathrm{dom}(\psi)$: $\Delta_i(e\mid \psi) \geq 0$.

\subsection{Adaptive Submodular Meta-Learning}
\label{sec:111}
We next formally introduce the adaptive submodular meta-learning framework.  Our framework is done in two stages: In the first (training) stage, we select a set $S\subseteq E$ of  size $l$ as the initial solution set, then, in the second (test) stage, we adaptively add the remaining $k-l$ items to $S$   after observing the  task at hand. The motivation behind pre-computing an initial solution $S$ is twofold: 1) in some cases it is time-consuming to acquire the partial realization of an item's state, it is more time-effective to acquire the partial realization of a batch of items' states at once. As in our case, the selection of $S$ is pre-computed regardless of the realization, we can safely select all items from $S$ at once and acquire their partial realization simultaneously, which significantly reduces the response time at the test time compared to full adaptive approach, i.e., $l=0$. 2) our second motivation inherits from the one behind the non-adaptive submodular meta-learning framework \citep{adibi2020submodular}. Since computing $S$ does not consume any computational resource such as power in the test stage, our framework helps to reduce the resource consumption in the test stage. We next consider two extreme cases in terms of $l$:

$\bullet$ $l=0$ means that we select all $k$ items adaptively after observing the incoming task. Clearly, this fully adaptive policy, which computes a fully personalized  solution for each incoming task, can not perform worse than  the case when $l>0$. However, implementing  such a fully adaptive policy is  more expensive than implementing a non-adaptive solution where all $k$ items are selected offline at once. For example, if acquiring the partial realization of an item's state is time-consuming, it is clearly more cost-effective to acquire the partial realization of a batch of items' states at once.

$\bullet$ $l=k$ means that we select all $k$ items offline before observing the incoming task. This non-adaptive policy is less computational power- and time-consuming since  it requires zero computation at the test time, however, because such a non-adaptive solution can not adapt to the incoming task, its performance could be less satisfactory than the one of an adaptive policy.

One important question that arises from the above discussion is what would be the best $l$. The answer to this question is application-specific, one can balance the computational overhead at test time  and the degree of personalization of the solution by tuning the value of $l$. We leave that decision to the decision-maker, and in this paper, we focus on finding the best learning policy for a given $l$.

\subsection{Problem Statement}
We use $\Omega(S, k)$ to denote the set of all policies that 1) pick $S$ as the initial set, i.e., for each $\pi\in \Omega(S, k)$ and each $i\in M$, it holds that $\pi(\emptyset, i)= S$, and 2)  $|E(\pi, \phi, i)|\leq k$ for all $\phi$ with $p(\phi)> 0$. Hence, the expected utility  $f^i_{avg}(\pi)$  of any $\pi \in \Omega(S, k)$ conditioned on observing task $i$ is \begin{eqnarray*}
f^i_{avg}(\pi)= f^i(S) + \Delta_i(\pi | S)
\end{eqnarray*}

The expected utility $\mathbb{E}_{i\sim \theta}[f^i_{avg}(\pi)]$ 
of a policy $\pi$ over the distribution $\theta$ of tasks can be written as
\begin{eqnarray*}
\mathbb{E}_{i\sim \theta}[f^i_{avg}(\pi)] =  \mathbb{E}_{i\sim \theta}[f^i(S)] + \mathbb{E}_{i\sim \theta}[\Delta_i(\pi | S)]
\end{eqnarray*}

Since the underlying probability distribution $\theta$ of the tasks is not always available,  we often collect a group of  tasks that are sampled  independently according to the distribution $\theta$. As a result, we focus on optimizing the sample
average approximation of $\mathbb{E}_{i\sim \theta}[f^i_{avg}(\pi)]$ given by
\begin{eqnarray*}
f_{avg}(\pi) &=& \frac{1}{m}\sum_{i\in M} f^i_{avg}(\pi)
\end{eqnarray*}

Thus, our problem can be written as
\begin{eqnarray}
\label{pb:1}
\max_{S\subseteq E, |S|=l} \max_{\pi\in \Omega(S, k)} f_{avg}(\pi)
\end{eqnarray}


\textbf{Remark 1:} It was worth noting that Problem (\ref{pb:1}) will be solved at training time to obtain a task-independent set $S$ of size $l$. The remaining $k-l$ items are added after observing the incoming task to form a task-specific solution of size $k$. When $l=0$, i.e., all $k$ items are selected adaptively after observing the incoming task,  our problem is reduced to the adaptive submodular maximization problem. This is because if the incoming task, say $i$, is known, our objective is reduced to maximizing $f^i_{avg}(\pi)$ subject to $|E(\pi, \phi, i)|\leq k$ for all $\phi$ with $p(\phi)> 0$. \cite{golovin2011adaptive} show that the following simple adaptive greedy algorithm achieves a $1-1/e$ approximation ratio for the monotone case. It starts with a empty set, and in each round, it selects an item that maximizes the marginal utility on top of the current partial realization. For the non-monotone case, \cite{tang2021beyond} develop a randomized policy that achieves a $1/e$ approximation ratio. Their policy starts with an empty set, and in each round, it selects an item uniformly at random from a set of $k$ items that have the largest marginal utility on top of the current partial realization. When $l = k$, i.e., all $k$ items are selected non-adaptively before observing the incoming task, our problem is reduced to the nonadaptive submodular maximization problem which has been studied in \citep{nemhauser1978analysis,buchbinder2014submodular}. Hence, in the rest of this paper, we assume that $l>0$ and $k-l>0$.

\textbf{Remark 2:} We next show that the training objective in Problem (\ref{pb:1})  is not adaptive submodular. Recall that we must select the first $l$ items before observing the incoming task. Because of the uncertainty associated with the incoming task, our utility function no longer satisfies the adaptive submodularity. Consider a toy example with two items $E=\{1, 2\}$, two tasks $M=\{1, 2\}$, $l=1$, and one state $O=\{1\}$, i.e., the state of each item is deterministic, and the utility functions are defined as follows: $f^1(\{1\}, 1)=f^1(\{2\}, 1)=f^1(\{1, 2\}, 1)=0$ and $f^2(\{1\}, 1)=f^2(\{2\}, 1)=1, f^2(\{1, 2\}, 1)=2$.  Hence, the marginal utility of item $1$ to an empty set before observing the task is $\frac{1}{2}\times f^1(\{1\}, 1) + \frac{1}{2}\times f^2(\{1\}, 1)=\frac{1}{2}$, meanwhile,  the marginal utility of item $1$ to an existing set $\{2\}$ after observing the incoming task $2$ is  $f^2(\{1, 2\}, 1) - f^2(\{2\}, 1)=1$, which is larger than $\frac{1}{2}$. This clearly violates the property of adaptive submodularity defined in (\ref{def:22}), making the existing results \cite{chen2013near} not applicable to our setting.

\section{Two-phase Greedy Policy for Monotone Case}
 We  first study the case when $f^i$ is adaptive submodular and adaptive monotone for all $i\in G$. We develop a \emph{Two-phase Greedy policy} $\pi^{g}$ to this case.  $\pi^{g}$ is composed of two phases: \emph{initialization phase} and \emph{execution phase}. The initialization phase is done at the training stage to find a good initial set $S^g$ of size $l$, the execution phase is conducted after observing the incoming task. 
 A detailed description of $\pi^{g}$ is listed in Algorithm \ref{alg:LPP1}.

$\bullet$ \textbf{Initialization Phase:}  Computing a task-independent initial set $S^g$ of size $l$ according to the follow classic non-adaptive greedy algorithm: It starts with $S^g=\emptyset$, and then adds a group of $l$ items to $S^g$ iteratively. In each step, it adds to $S^g$ an item that maximizes the marginal utility of the average approximation of $m$ utility functions on top of the selected items. 
  This process iterates until all $l$ items have been added to $S^g$.

   $\bullet$ \textbf{Execution Phase:}  When a task $i\in G$ arrives, $\pi^{g}$ first selects $S^g$ and observe their states, then selects the rest of the $k-l$ items according to an adaptive greedy algorithm. In particular,  $\pi^{g}$ runs in $k$ rounds and it selects one item in each round. The first $l$ rounds are performed non-adaptively for selecting $S^g$ and observing  the partial realization $\psi^{g_l}$ of  $S^g$. The remaining $k-l$ rounds are performed adaptively for selecting the rest $k-l$ items as follows: At each of the remaining $k-l$ rounds $t\in[l+1, k]$, it selects an item that maximizes the expected marginal utility of $f^i$ on top of the current partial realization $\psi^{g_{t-1}}$:
       \[e_t\leftarrow \arg\max_{e \in E}\Delta_i(e\mid \psi^{g_{t-1}})\]
    After observing the state $\phi(e_t)$ of $e_t$, update  the current partial realization $\psi^{g_t}$ using  $\psi^{g_{t-1}}\cup\{\phi(e_t)\}$.  This process iterates until all the remaining $k-l$ items have been selected.

\begin{algorithm}[hptb]
\caption{Two-phase  Greedy Policy $\pi^{g}$}
\label{alg:LPP1}
\begin{algorithmic}[1]
\STATE $S^g=\emptyset, t=1, b=1, \psi^{g_{0}}=\emptyset$.

\COMMENT {\underline{Initialization Phase}}
\WHILE {$b \leq l$}
\STATE $S^g \leftarrow S^g\cup \arg\max_{e \in E}\frac{1}{m}\sum_{i\in M} (f^i(S^g\cup\{e\})-f^i(S^g))$; $b\leftarrow b+1$;
\ENDWHILE

\COMMENT {\underline{Execution Phase}}

\COMMENT {The first $l$ rounds are performed non-adaptively for selecting $S^g$.}
\FOR {$e\in S^g$}
\STATE  $e_t\leftarrow e$;
\STATE select $e_t$ and observe $\phi(e_t)$;  $\psi^{g_t} = \psi^{g_{t-1}}\cup\{\phi(e_t)\}$;  $t\leftarrow t+1$;
\ENDFOR

\COMMENT {The remaining  $k-l$ rounds are performed adaptively.}
\WHILE {$t \leq k$}
\STATE observe $\psi^{g_{t-1}}$;
\STATE $e_t\leftarrow \arg\max_{e \in E}\Delta_i(e\mid  \psi^{g_{t-1}})$;
\STATE select $e_t$ and observe $\phi(e_t)$;
\STATE $\psi^{g_t} = \psi^{g_{t-1}}\cup\{\phi(e_t)\}$;  $t\leftarrow t+1$;
\ENDWHILE
\end{algorithmic}
\end{algorithm}

The rest of this section is devoted to proving the performance bound of $\pi^{g}$. We use $\pi^{o}$ to denote an optimal policy and use $S^{o}$ to denote the initial set selected by  $\pi^{o}$.  Here, we assume that $S^o$ is deterministic since a probabilistic initial solution set  can be expressed as the weighted sum of deterministic initial solution sets. Before presenting the main theorem (Theorem \ref{thm:1}) of this paper, we first present three preparatory lemmas.
\begin{lemma}
\label{lem:1}For all $i\in M$, we have
$f^i_{avg}(\pi^{g}@ \pi^{o})- f^i_{avg}(\pi^{g} @ \pi^{o}_l) \leq \Delta_i(\pi^{g} | S^g)$, where $ \pi^{o}_l$ denotes the level-$l$-truncation of $\pi^{o}$.
\end{lemma}
\emph{Proof:} Let $\psi^{g@o_{t'}}=\psi^{g}\cup \psi^{o_{t'}}$ denote the partial realization obtained after running $\pi^{g} @\pi^{o}_{t'}$, where $\psi^{g}$ is the partial realization after running $\pi^{g}$, and $\psi^{o_{t'}}$ is the partial realization after running $\pi^{o}_{t'}$. In addition, we use $\psi^{g_{t''}} \subseteq \psi^{g@o_{t'}}$ to denote the partial realization after running the level-$t''$-truncation of $\pi^{g}$. Let random variable $e^{opt}_{t}$ denote the $t$-th item selected by $\pi^{o}$ conditioned on the current partial realization  $\psi^{o_{t-1}}$, we first bound the expected marginal utility $\Delta_i(e_t\mid \psi^{g_{t-1}})$ of $e _{t}$ for any $t\in[l+1, k]$ and any  $\psi^{g_{t-1}}$. For each $t\in[l+1, k]$, we have
\begin{eqnarray}
&&\Delta_i(e _{t} \mid \psi^{g_{t-1}})= \max_{e\in E}\Delta_i(e \mid  \psi^{g_{t-1}})~\nonumber\\
&&\geq \max_{e\in E}\Delta_i(e \mid \psi^{g@o_{t-1}})\geq \mathbb{E}_{\Pi^o}[\Delta_i(e^{opt}_{t} \mid \psi^{g@o_{t-1}})]~\nonumber\\
&&= f_{avg}(\pi^{g} @\pi^{o}_{t}\mid \psi^{g@o_{t-1}})-f_{avg}(\pi^{g} @\pi^{o}_{t-1}\mid \psi^{g@o_{t-1}})~\nonumber
\end{eqnarray}
The first equality is due to $\pi^{g}$ selects an item that maximizes the conditional expected marginal utility conditioned on $\psi^{g_{t-1}}$. The first inequality is due to the assumption that $f^i$ is adaptive submodular and $\psi^{g_{t-1}}\subseteq  \psi^{g@o_{t-1}}$.

Unfixing $\psi^{g@o_{t-1}}$ and take the expectation over $(\Psi^{g}, \Psi^{o}_t)$, the following inequality holds for all $t\in[l+1,k]$:
$\mathbb{E}_{\Psi^{g@o_{t-1}}}[\Delta_i(e _{t} \mid \Psi^{g_{t-1}})] \geq \mathbb{E}_{\Psi^{g@o_{t-1}}}[f_{avg}(\pi^{g} @\pi^{o}_{t}\mid \Psi^{g@o_{t-1}})-f_{avg}(\pi^{g} @\pi^{o}_{t-1}\mid \Psi^{g@o_{t-1}})]$. It follows that
\begin{eqnarray}
f_{avg}(\pi^{g}_t)-f_{avg}(\pi^{g}_{t-1})\geq f_{avg}(\pi^{g} @\pi^{o}_{t})-f_{avg}(\pi^{g} @\pi^{o}_{t-1})~\nonumber
\end{eqnarray}
We further have
\begin{eqnarray}
&&\sum_{t\in[l+1, k]}(f_{avg}(\pi^{g}_t)-f_{avg}(\pi^{g}_{t-1}))~\nonumber \\
&&\geq \sum_{t\in[l+1, k]}(f_{avg}(\pi^{g} @\pi^{o}_{t})-f_{avg}(\pi^{g} @\pi^{o}_{t-1}))
\label{eq:aaa}
\end{eqnarray}
This lemma holds due to $\Delta_i(\pi^{g} | S^g) = \sum_{t\in[l+1, k]}(f_{avg}(\pi^{g}_t)-f_{avg}(\pi^{g}_{t-1}))$ and (\ref{eq:aaa}). $\Box$

\begin{lemma}
\label{lem:2}For all $i\in M$, we have $\Delta_i(\pi^{g} | S^{g})\geq \Delta_i(\pi^{g} | S^{g}\cup S^{o})$.
\end{lemma}
\emph{Proof:} We first bound the expected marginal utility $\Delta_i(e_t\mid \psi^{g_{t-1}})$ of $e _{t}$ conditioned on partial realization $\psi^{g_{t-1}}$ for all $t\in[k]$: $\Delta_i(e_t\mid \psi^{g_{t-1}})\geq \mathbb{E}[\Delta_i(e_t\mid \Phi(S^{o})\cup \psi^{g_{t-1}})\mid \Phi \sim \psi^{g_{t-1}}]$. This inequality is due to $\psi^{g_{t-1}} \subseteq \Phi(S^{o})\cup \psi^{g_{t-1}}$ for any $\Phi$ such that $\Phi \sim \psi^{g_{t-1}}$, and $f^i$ is adaptive submodular. 

Unfixing $\psi^{g_{t-1}}$ and take the expectation over $\Psi^{g_{t-1}}$, we have $\mathbb{E}_{\Psi^{g_{t-1}}}[\Delta_i(e_t\mid \Psi^{g_{t-1}})]\geq \mathbb{E}_{\Psi^{g_{t-1}}}[ \mathbb{E}[\Delta_i(e_t\mid \Phi(S^{o})\cup \Psi^{g_{t-1}})\mid \Phi \sim \Psi^{g_{t-1}}]]$. Hence, the following inequality holds for all $t\in[k]$:
{\small\begin{eqnarray}\label{eq:important}
f_{avg}(\pi^{g}_t)- f_{avg}(\pi^{g}_{t-1})\geq f_{avg}(\pi^{g}_t @ \pi^{o}_l)-f_{avg}(\pi^{g}_{t-1} @ \pi^{o}_l)
\end{eqnarray}}

Then we have $
\Delta_i(\pi^{g} | S^{g}) = \sum_{t=l+1}^{k} (f_{avg}(\pi^{g}_t)- f_{avg}(\pi^{g}_{t-1}))
\geq \sum_{t=l+1}^{k} (f_{avg}(\pi^{g}_t @ \pi^{o}_l)-f_{avg}(\pi^{g}_{t-1} @ \pi^{o}_l))=  \Delta_i(\pi^{g} | S^{g}\cup S^{o})$. The first inequality is due to (\ref{eq:important}). This finishes the proof of this lemma. $\Box$

\begin{lemma}
\label{lem:3}
$\sum_{i\in M}(f^i(S^g\cup S^o)-f^i(S^g)) \leq  \sum_{i\in M}f^i(S^g)$.
\end{lemma}
\emph{Proof:} Because $f^i$ is adaptive monotone and adaptive submodular for all $i\in M$, $f^i(S) = \mathbb{E}_{\Phi\sim p}[f^i(S, \Phi)]$ is monotone and submodular in terms of $S$. Hence,  $\sum_{i\in M}f^i(S)$ is also monotone and submodular in terms of $S$ due the linear combination of monotone submodular functions are still monotone and submodular.  
 Because we apply the classic non-adaptive greedy algorithm \cite{fisher1978analysis} to obtain $S^g$, this lemma holds due to the same analysis of Theorem 2.1 in \cite{fisher1978analysis}. $\Box$ 

Now we are ready to present the first main theorem of this paper.

\begin{theorem} \label{thm:1}Our two-phase greedy policy $\pi^g$ achieves a $1/2$ approximation ratio, that is,
$
f_{avg}(\pi^{g}) \geq \frac{1}{2} f_{avg}(\pi^{o})$.
\end{theorem}

%
%

\emph{Proof:} Recall that $\pi^{g}@ \pi^{o}$ runs $\pi^{g}$ first, then runs  $\pi^{o}$ from a fresh start. Hence, the expected utility $f^i_{avg}(\pi^{g}@ \pi^{o})$ of $\pi^{g}@ \pi^{o}$ from task $i$ can be written as:
\begin{eqnarray*}&&f^i_{avg}(\pi^{g}@ \pi^{o})= f^i(S^g) + (f^i(S^g\cup S^o)-f^i(S^g))+ \Delta_i(\pi^{g} | S^g\cup S^o) +(f^i_{avg}(\pi^{g}@ \pi^{o})- f^i_{avg}(\pi^{g} @ \pi^{o}_l))
\end{eqnarray*}

It follows that
\begin{eqnarray*}
&& m\times f_{avg}(\pi^{g}@ \pi^{o}) = \sum_{i\in M} f^i_{avg}(\pi^{g}@ \pi^{o})\\
&&=   \sum_{i\in M}f^i(S^g) +  \sum_{i\in M}(f^i(S^g\cup S^o)-f^i(S^g))+  \sum_{i\in M}\Delta_i(\pi^{g} | S^g\cup S^o)   +  \sum_{i\in M}(f^i_{avg}(\pi^{g}@ \pi^{o})- f^i_{avg}(\pi^{g} @ \pi^{o}_l) )\\
   &&\leq 2 \sum_{i\in M}f^i(S^g)+ \sum_{i\in M}\Delta_i(\pi^{g} | S^g\cup S^o)+  \sum_{i\in M}(f^i_{avg}(\pi^{g}@ \pi^{o})- f^i_{avg}(\pi^{g} @ \pi^{o}_l) )\\
    &&\leq 2\sum_{i\in M}f^i(S^g)+   \sum_{i\in M}\Delta_i(\pi^{g} | S^g)
+  \sum_{i\in M}\Delta_i(\pi^{g} | S^g)\\
    &&=  2(\sum_{i\in M}f^i(S^g)+  \sum_{i\in M}\Delta_i(\pi^{g} | S^g))=  2   \sum_{i\in M}f^i_{avg}(\pi^{g})=m\times 2f_{avg}(\pi^{g})
\end{eqnarray*}
The first inequality is due to Lemma \ref{lem:3}, the second inequality is due to Lemma \ref{lem:1} and Lemma \ref{lem:2}. It follows that $f_{avg}(\pi^{g}) \geq \frac{1}{2} f_{avg}(\pi^{g}@ \pi^{o})$. Together with $f_{avg}(\pi^{g}@ \pi^{o})\geq  f_{avg}(\pi^{o})$ due to $f^i$ is adaptive monotone for all $i\in M$, we have $f_{avg}(\pi^{g}) \geq \frac{1}{2} f_{avg}(\pi^{o})$.  $\Box$

\section{Two-phase Randomized Greedy Policy for Non-Monotone Case}
 We next explain the design of our \emph{Two-phase Randomized Greedy policy} $\pi^{r}$ for the non-monotone adaptive meta-learning problem.  Similar to the solution designed for the monotone case, $\pi^{r}$ is composed of two phases: \emph{initialization phase} and \emph{execution phase}. The initialization phase is done at the training stage to find a good initial set $S^r$ of size $l$, the execution phase is conducted after observing the incoming task. 
 We first add a set $D$ of $k$ dummy items to the ground set, such that, for any $e \in D$, any
partial realization $\psi$, and any $i\in G$, we have $\Delta_i(d \mid \psi) =0$. Let $E'=E\cup D$. We introduce $D$ to ensure that our solution  never adds an item with negative marginal utility to the solution. Note that we can safely remove those dummy items from the solution without affecting its utility. A detailed description of $\pi^{r}$ is listed in Algorithm \ref{alg:LPP2}.

$\bullet$ \textbf{Initialization Phase:}  Computing a task-independent initial set $S^r$ of size $l$ according to the following  non-adaptive random greedy algorithm: It starts with $S^r=\emptyset$, and then adds a group of $l$ items to $S^r$ iteratively: At each round $b \in[l]$, we  select an
item uniformly at random from the set $U(S^r)$, which contains the $l$ items with the largest marginal utility to the current solution $S^r$, and add it to $S^r$. This process iterates until all $l$ items are added to $S^r$.

$\bullet$ \textbf{Execution Phase:}  
$\pi^r$ runs in $k$ rounds. The first $l$ rounds are performed non-adaptively for selecting $S^r$ and observing  the partial realization $\psi^{r_l}$ of  $S^r$. The remaining $k-l$ rounds are performed adaptively for selecting the rest $k-l$ items after observing the incoming task, say $i\in G$: At each of the remaining $k-l$ rounds $t\in[l+1, k]$,  $\pi^r$  selects an
item $e_t$ uniformly at random from the set $U(\psi^{r_{t-1}})$, where $U(\psi^{r_{t-1}})$ contains the $k-l$ items with the largest marginal utility on top of the current partial realization $\psi^{r_{t-1}}$.
    After observing the state $\phi(e_t)$ of $e_t$, update  the current partial realization $\psi^{r_t}$ using  $\psi^{r_{t-1}}\cup\{\phi(e_t)\}$.  This process iterates until all the remaining $k-l$ items have been selected.

\begin{algorithm}[hptb]
\caption{Two-phase Randomized Greedy policy $\pi^r$}
\label{alg:LPP2}
\begin{algorithmic}[1]
\STATE $S^r=\emptyset, t=1, b=1, \psi^{r_{0}}=\emptyset$.

\COMMENT {\underline{Initialization Phase}}
\WHILE {$b \leq l$}
\STATE $U(S^r) \leftarrow \arg\max_{V \in E': |V|\leq l}\frac{1}{m}\sum_{e\in V}\sum_{i\in M} (f^i(S^r\cup\{e\})-f^i(S^r))$;
\STATE sample $e$ uniformly at random from $U(S^r)$;
\STATE $S^r\leftarrow S^r\cup\{e\}$
\STATE $b\leftarrow b+1$;
\ENDWHILE

\COMMENT {\underline{Execution Phase}}

\COMMENT {The first $l$ rounds are performed non-adaptively for selecting $S^r$.}
\FOR {$e\in S^r$}
\STATE  $e_t\leftarrow e$;
\STATE select $e_t$ and observe $\phi(e_t)$;  $\psi^{r_t} = \psi^{r_{t-1}}\cup\{\phi(e_t)\}$;  $t\leftarrow t+1$;
\ENDFOR

\COMMENT {The remaining  $k-l$ rounds are performed adaptively.}
\WHILE {$t \leq k$}
\STATE observe $\psi^{r_{t-1}}$;
\STATE $U(\psi^{r_{t-1}}) \leftarrow \arg\max_{V \in E': |V|\leq k-l}\sum_{e\in V} \Delta_i(e\mid  \psi^{r_{t-1}})$;
\STATE sample $e$ uniformly at random from $U(\psi^{r_{t-1}})$;
\STATE  $e_t\leftarrow e$;
\STATE select $e_t$ and observe $\phi(e_t)$;
\STATE $\psi^{r_t} = \psi^{r_{t-1}}\cup\{\phi(e_t)\}$;  $t\leftarrow t+1$;
\ENDWHILE
\end{algorithmic}
\end{algorithm}

The rest of this section is devoted to proving the performance bound of $\pi^{g}$. We still use $\pi^o$ to denote the optimal policy and use $S^o$ to denote the initial solution set adopted by $\pi^o$. Before presenting the main theorem, we first present four technical lemmas.

\begin{lemma}
\label{lem:1b}For all $i\in M$, we have
$f^i_{avg}(\pi^r@ \pi^{o})- f^i_{avg}(\pi^r @ \pi^{o}_l) \leq f^i_{avg}(\pi^r)- f^i_{avg}(\pi^r_l)$, where $ \pi^{o}_l$ (resp. $\pi^r_l$) denotes the level-$l$-truncation of $\pi^{o}$ (resp. $\pi^r$).
\end{lemma}
\emph{Proof:} Define $\psi^{r_{0}}=\emptyset$, let $\overrightarrow{\psi^r}=\{\psi^{r_{0}}, \psi^{r_{1}}, \psi^{r_{2}},\cdots, \psi^{r_{k}}\}$ denote the sequence of partial realizations obtained after running $\pi^r$, where $\psi^{r_{t}}$ is the partial realization observed after selecting the $t$-th item. Conditioned a sequence of partial realizations $\overrightarrow{\psi^r}$, we first give an upper bound on the value of $f^i_{avg}(\pi^r@ \pi^{o})- f^i_{avg}(\pi^r @ \pi^{o}_l)$. Let $A_e$ be an indicator that $e$ is selected by $\pi^o$ at some step $t$ such that $t>l$, and  let $x_e$ denote the expected marginal contribution of $e$ to $f^i_{avg}(\pi^r@ \pi^{o})- f^i_{avg}(\pi^r @ \pi^{o}_l)$ conditioned on $A_e=1$ and a partial realization $\psi^{r_{k}}$.
\begin{eqnarray*}
\mathbb{E}[f^i_{avg}(\pi^r@ \pi^{o})- f^i_{avg}(\pi^r @ \pi^{o}_l)\mid \overrightarrow{\psi^r}]&=&\sum_{e\in E'} \Pr[A_e=1 \mid \psi^{r_{k}}]\cdot x_e\\
&\leq&  \sum_{e\in E'}  \Pr[A_e=1 \mid \psi^{r_{k}}]\cdot \Delta_i(e\mid \psi^{r_{k}})\\
&\leq& \max_{V: |V|\leq k-l}\sum_{e\in V} \Delta_i(e\mid \psi^{r_{k}})
\end{eqnarray*}
 Consider an arbitrary item $e\in E'$, assume $e$ is selected by $\pi^o$ at some step $t$ such that $t>l$, the first inequality is due to $f^i$ is adaptive submodular and $\psi^{r_{k}}$ is a subrealization of the realization observed after running $\pi^r @ \pi^{o}_{t-1}$. The second inequality is due to $\sum_{e\in E'} \Pr[A_e=1 \mid \psi^{r_{k}}]\leq k-l$ for all  $\psi^{r_{k}}$, this is because $\pi^o$ selects at most $k-l$ items at the test stage. Hence,
\begin{eqnarray}
f^i_{avg}(\pi^r@ \pi^{o})- f^i_{avg}(\pi^r @ \pi^{o}_l)&=& \sum_{\overrightarrow{\psi^r}}\Pr[\overrightarrow{\psi^r}] \mathbb{E}[f^i_{avg}(\pi^r@ \pi^{o})- f^i_{avg}(\pi^r @ \pi^{o}_l)\mid \overrightarrow{\psi^r}]~\nonumber\\
&\leq&  \sum_{\overrightarrow{\psi^r}}\Pr[\overrightarrow{\psi^r}]  \max_{V: |V|\leq k-l}\sum_{e\in V} \Delta_i(e\mid \psi^{r_{k}})\label{eq:panther}
\end{eqnarray}
where $\Pr[\overrightarrow{\psi^r}]$ denotes the probability that $\overrightarrow{\psi^r}$ is realized. We next provide a lower bound of $f^i_{avg}(\pi^r)- f^i_{avg}(\pi^r_l)$.
\begin{eqnarray}
f^i_{avg}(\pi^r)- f^i_{avg}(\pi^r_l) &=&  \sum_{\overrightarrow{\psi^r}} \Pr[\overrightarrow{\psi^r}]\sum_{t\in[l+1, k]} \sum_{e\in U(\psi^{r_{k}})}\frac{1}{k}\Delta_i(e\mid \psi^{r_{t}})\\
&\geq& \sum_{\overrightarrow{\psi^r}} \Pr[\overrightarrow{\psi^r}]\sum_{t\in[l+1, k]} \sum_{e\in U(\psi^{r_{k}})}\frac{1}{k}\Delta_i(e\mid \psi^{r_{k}})~\nonumber\\
&=& \sum_{\overrightarrow{\psi^r}} \Pr[\overrightarrow{\psi^r}] \sum_{e\in U(\psi^{r_{k}})}\Delta_i(e\mid \psi^{r_{k}})~\nonumber\\
&=&  \sum_{\overrightarrow{\psi^r}} \Pr[\overrightarrow{\psi^r}]  \max_{V: |V|\leq k-l}\sum_{e\in V} \Delta_i(e\mid \psi^{r_{k}}) \label{eq:panther2}
\end{eqnarray}
The first equality is due to the design of $\pi^r$, i.e., at each round $t\in[l+1, k]$, $\pi^r$ selects an item $e_t$ uniformly at random from the set $U(\psi^{r_{t-1}})$. The first inequality is due to $f^i$ is adaptive submodular and $\psi^{r_{k}}\supseteq  \psi^{r_{t}}$ for all $t\in[k]$. The third equality is due to definition of $U(\psi^{r_{t-1}})$, i.e.,  $U(\psi^{r_{t-1}}) = \arg\max_{V \in E': |V|\leq k-l}\sum_{e\in V} \Delta_i(e\mid  \psi^{r_{t-1}})$. (\ref{eq:panther}) and (\ref{eq:panther2}) together imply that $f^i_{avg}(\pi^r@ \pi^{o})- f^i_{avg}(\pi^r @ \pi^{o}_l) \leq f^i_{avg}(\pi^r)- f^i_{avg}(\pi^r_l)$. This finishes the proof of this lemma. $\Box$

The following lemma can be proved by following the same proof of Lemma \ref{lem:2}.
\begin{lemma}
\label{lem:2b}
For all $i\in M$, we have $f^i_{avg}(\pi^o_l@ \pi^r)-f^i_{avg}(\pi^o_l@ \pi_l^r)\leq f^i_{avg}(\pi^r)-f^i_{avg}(\pi_l^r)$.
\end{lemma}

\begin{lemma}
\label{lem:3b}
$\sum_{i\in M}(f^i_{avg}(\pi^r_l@ \pi^o_l)-f^i_{avg}(\pi^r_l)) \leq  \sum_{i\in M}f^i_{avg}(\pi^r_l)$.
\end{lemma}
\emph{Proof:}  Define $g(S)=\sum_{i\in M}f^i(S)$. Because $f^i$ is adaptive submodular for all $i\in M$, $f^i(S) = \mathbb{E}_{\Phi\sim p}[f^i(S, \Phi)]$ is submodular in terms of $S$. Hence, $g(S)$ is also submodular in terms of $S$ due the linear combination of  submodular functions is still  submodular.   Because the first $l$ items are selected non-adaptively by both $\pi^r$ and $\pi^o$, proving this lemma is equivalent to show that  $\mathbb{E}_{S^r}[g(S^r\cup S^o)-g(S^r)] \leq  \mathbb{E}_{S^r}[g(S^r)]$.  Define $S^{r_{0}}=\emptyset$, let $\overrightarrow{S^r}=\{S^{r_{0}}, S^{r_{1}}, S^{r_{2}},\cdots, S^{r_{l}}\}$ denote the sequence of sets selected by $\pi^r$, where $ S^{r_{t}}$ denotes the first $t$ items selected by $\pi^r$. Sort $S^o$ in an arbitrary order, for each $t\in [l]$, let $S^{o}[t]$ denote the $t$-th item of $S^o$. We first provide an upper bound of $\mathbb{E}_{S^r}[g(S^r\cup S^o)-g(S^r)]$.
\begin{eqnarray}
\mathbb{E}_{S^r}[g(S^r\cup S^o)-g(S^r)] &=&\mathbb{E}_{S^r}[ \sum_{t\in [l]} ( g(S^{o}[t]\cup S^r)- g(S^{o}[t-1]\cup S^r))]~\nonumber\\
&\leq&  \mathbb{E}_{S^r}[ \sum_{e\in S^o} (g(\{e\}\cup S^r)- g(S^r))]~\nonumber\\
&\leq& \mathbb{E}_{S^r}[ \max_{V: |V|\leq l}\sum_{e\in V} (g(\{e\}\cup S^r)- g(S^r))] \label{eq:ch}
\end{eqnarray}
The first inequality is due to $f^i$ is adaptive submodular. We next provide a lower bound of $\mathbb{E}_{S^r}[g(S^r)]$.
\begin{eqnarray}
\mathbb{E}_{S^r}[g(S^r)] &=&\mathbb{E}_{S^r}[ \sum_{t\in [l]} \frac{1}{l} \sum_{e\in U(S^{r_{t-1}})}(g(S^{r_{t-1}}\cup \{e\})- g(S^{r_{t-1}}))]~\nonumber\\
&=& \mathbb{E}_{S^r}[ \sum_{t\in [l]} \frac{1}{l} \max_{V: |V|\leq l}\sum_{e\in V} (g(\{e\}\cup S^{r_{t-1}})- g(S^{r_{t-1}}))]~\nonumber\\
&\geq& \mathbb{E}_{S^r}[ \sum_{t\in [l]} \frac{1}{l} \max_{V: |V|\leq l}\sum_{e\in V} (g(\{e\}\cup S^r)- g(S^r))]~\nonumber\\
&=& \mathbb{E}_{S^r}[ \max_{V: |V|\leq l}\sum_{e\in V} (g(\{e\}\cup S^r)- g(S^r))]\label{eq:ch1}
\end{eqnarray}
The first equality is due to the design of $\pi^r$, i.e., it selects an item uniformly at random from $U(S^{r_{t-1}})$, the inequality is due to $g$ is submodular and $S^{r_{t-1}} \subseteq S^r$ for all $t\in[l]$. (\ref{eq:ch}) together with (\ref{eq:ch1}) imply that $\mathbb{E}_{S^r}[g(S^r\cup S^o)-g(S^r)] \leq  \mathbb{E}_{S^r}[g(S^r)]$. This finishes the proof of this lemma. $\Box$

\begin{lemma}
\label{qqq}
$f_{avg}(\pi^r@\pi^o)\geq (1-\frac{1}{l})^l(1-\frac{1}{k-l})^{k-l}f_{avg}(\pi^o)$.
\end{lemma}
\emph{Proof:} Recall that in each of the first $l$ rounds of $\pi^r$, it selects an item uniformly at random from a set $U(S^r)$ of $l$ items. According to Lemma 1 in \citep{tang2021beyond}, if $f^i$ is adaptive submodular, we have
$f^i_{avg}(\pi^r_l@\pi^o)\geq (1-\frac{1}{l})^l f^i_{avg}(\pi^o)$. Similarly, since in each of the last $k-l$ rounds of $\pi^r$, it selects an item randomly from a set $U(\psi^{r_{t-1}})$ of $k-l$ items, then we have $f^i_{avg}(\pi^r@\pi^o)\geq (1-\frac{1}{k-l})^{k-l}f^i_{avg}(\pi^r_l@\pi^o)$ if $f^i$ is adaptive submodular. It follows that $f^i_{avg}(\pi^r@\pi^o)\geq (1-\frac{1}{l})^l(1-\frac{1}{k-l})^{k-l}f^i_{avg}(\pi^o)$. Because $f_{avg}(\pi^r@\pi^o)=\frac{1}{m}\sum_{i\in M} f^i_{avg}(\pi^r@\pi^o)$ and $f_{avg}(\pi^o)=\frac{1}{m}\sum_{i\in M} f^i_{avg}(\pi^o)$, we have $f_{avg}(\pi^r@\pi^o)\geq (1-\frac{1}{l})^l(1-\frac{1}{k-l})^{k-l}f_{avg}(\pi^o)$. $\Box$

Now we are ready to present the second main theorem of this paper.

\begin{theorem} \label{thm:1b}Our two-phase randomized greedy policy $\pi^r$ achieves a $\frac{1}{2} (1-\frac{1}{l})^l(1-\frac{1}{k-l})^{k-l} $ approximation ratio, that is,
$
f_{avg}(\pi^r) \geq \frac{1}{2} (1-\frac{1}{l})^l(1-\frac{1}{k-l})^{k-l}  f_{avg}(\pi^{o})$.
\end{theorem}

%
%

\emph{Proof:} Recall that $\pi^r@ \pi^{o}$ runs $\pi^r$ first, then runs  $\pi^{o}$ from a fresh start. Hence, the expected utility $f^i_{avg}(\pi^r@ \pi^{o})$ of $\pi^r@ \pi^{o}$ from task $i$ can be written as:
\begin{eqnarray*}f^i_{avg}(\pi^r@ \pi^{o})&&= f^i_{avg}(\pi^r_l) + (f^i_{avg}(\pi^r_l@ \pi^{o}_l)-f^i_{avg}(\pi^r_l)) \\
&&+ (f^i_{avg}(\pi^o_l@ \pi^r)-f^i_{avg}(\pi^o_l@ \pi_l^r)) +(f^i_{avg}(\pi^r@ \pi^{o})- f^i_{avg}(\pi^r @ \pi^{o}_l))
\end{eqnarray*}

It follows that
{\small\begin{eqnarray*}
&& m\times f_{avg}(\pi^r@ \pi^{o}) = \sum_{i\in M} f^i_{avg}(\pi^r@ \pi^{o})\\
&&=    \sum_{i\in M}f^i_{avg}(\pi^r_l) +  \sum_{i\in M}(f^i_{avg}(\pi^r_l@ \pi^{o}_l)-f^i_{avg}(\pi^r_l)) +  \sum_{i\in M}(f^i_{avg}(\pi^o_l@ \pi^r)-f^i_{avg}(\pi^o_l@ \pi_l^r)) + \sum_{i\in M}(f^i_{avg}(\pi^r@ \pi^{o})- f^i_{avg}(\pi^r @ \pi^{o}_l))\\
&&\leq   2 \sum_{i\in M}f^i_{avg}(\pi^r_l) +  \sum_{i\in M}(f^i_{avg}(\pi^o_l@ \pi^r)-f^i_{avg}(\pi^o_l@ \pi_l^r)) + \sum_{i\in M}(f^i_{avg}(\pi^r@ \pi^{o})- f^i_{avg}(\pi^r @ \pi^{o}_l))\\
&&\leq   2 (\sum_{i\in M}f^i_{avg}(\pi^r_l) +   \sum_{i\in M}(f^i_{avg}(\pi^r)- f^i_{avg}(\pi^r_l)))=  2   \sum_{i\in M}f^i_{avg}(\pi^r)=m\times 2f_{avg}(\pi^r)
\end{eqnarray*}}
The first inequality is due to Lemma \ref{lem:3b}, the second inequality is due to Lemma \ref{lem:1b} and Lemma \ref{lem:2b}. It follows that $f_{avg}(\pi^r) \geq \frac{1}{2} f_{avg}(\pi^r@ \pi^{o})$. Together with $f_{avg}(\pi^r@\pi^o)\geq (1-\frac{1}{l})^l(1-\frac{1}{k-l})^{k-l}f_{avg}(\pi^o)$ due to Lemma \ref{qqq}, we have
$
f_{avg}(\pi^r) \geq \frac{1}{2} (1-\frac{1}{l})^l(1-\frac{1}{k-l})^{k-l} f_{avg}(\pi^{o})$. $\Box$

Theorem \ref{thm:1b}, together with the fact that $(1-\frac{1}{l})^l\geq 1/4$ and $(1-\frac{1}{k-l})^{k-l}\geq 1/4$ when $l>1$ and $k-l> 1$, implies the following corollary.

\begin{corollary}
When $l>1$ and $k-l> 1$, our two-phase randomized greedy policy $\pi^r$ achieves a $1/32$ approximation ratio, that is, $f_{avg}(\pi^r) \geq \frac{1}{32}   f_{avg}(\pi^{o})$.
\end{corollary}

We next discuss the remaining cases when  $l=1$ or $k-l=1$.

\paragraph{A $\frac{1}{1+e}$-approximate solution when $k-l=1$.} When $k-l=1$, the first $k-1$ items are selected non-adaptively in the training stage, and the last one item is selected after observing the incoming task. Our solution $\pi^a$ is to randomly pick a policy from  $\pi^{a1}$ and $\pi^{a2}$ to follow such that $\pi^{a1}$ is picked with probability $\frac{1}{1+e}$ and $\pi^{a2}$ is picked with probability $\frac{e}{1+e}$. We next describe the details of $\pi^{a1}$ and $\pi^{a2}$.
\begin{itemize}
\item The first candidate solution $\pi^{a1}$ is a non-adaptive solution, which selects a fixed set  $S^{a}$  of items of size $k-l$ for all incoming tasks. We compute $S^{a}$ using the greedy algorithm described in the Initialization phase of $\pi^r$.
\item The second candidate solution $\pi^{a2}$ does not select any items during the initialization phase, after observing the incoming task, say $i\in G$, it picks  a singleton $e(i)$ with the largest expected utility, i.e., $e(i)=\arg\max_{e\in E'} f^i(\{e\})$.
\end{itemize}

\begin{theorem}
When $k-l=1$, $\pi^{a}$ achieves a $\frac{1}{1+e}$ approximation ratio, i.e.,  $f_{avg}(\pi^{a}) \geq \frac{1}{1+e}f_{avg}(\pi^{o})$.
\end{theorem}
\emph{Proof:}
According to the design of $\pi^a$, it picks $\pi^{a1}$ (resp. $\pi^{a2}$) with probability $\frac{1}{1+e}$  (resp. $\frac{e}{1+e}$). Hence,  the expected utility $f^i_{avg}(\pi^{a})$ of $\pi^a$,  for any $i\in G$, can be derived as follows:
\begin{eqnarray}
f_{avg}(\pi^{a}) &=&  \frac{1}{1+e}\sum_{i\in M} f^i_{avg}(\pi^{a1})+ \frac{e}{1+e}\sum_{i\in M} f^i_{avg}(\pi^{a2})~\nonumber\\
&=& \frac{1}{m}\frac{1}{1+e}\sum_{i\in M} f^i(S^{a})+ \frac{1}{m}\frac{e}{1+e}\sum_{i\in M} f^i(\{e(i)\}) \label{eq:tb}
\end{eqnarray}
We next derive the expected utility $f^i_{avg}(\pi^{o})$ of $\pi^o$ for any $i\in G$.
\begin{eqnarray}
\label{eq:tb1}f^i_{avg}(\pi^{o})&&= f^i(S^o) + \mathbb{E}_{\Psi^{o_{k-1}}, \Pi^o}[\Delta_i( e(\pi^o,  \Psi^{o_{k-1}}, i) \mid \Psi^{o_{k-1}})\mid \Psi^{o_{k-1}}]
\end{eqnarray}
where $\Psi^{o_{k-1}}$ denotes a random realization of the states of $S^o$, and  $e(\pi^o,  \Psi^{o_{k-1}}, i)$ denotes the (random) item selected by $\pi^o$ after observing the incoming task $i\in G$ and partial realization $\Psi^{o_{k-1}}$.
Because $f^i$ is adaptive submodular, we have $\Delta_i( e(\pi^o,  \Psi^{o_{k-1}}, i) \mid \Psi^{o_{k-1}})\leq \Delta_i( e(\pi^o,  \Psi^{o_{k-1}}, i) \mid \emptyset)$ due to $\emptyset\subseteq\Psi^{o_{k-1}}$. Moreover, because $e(i)=\arg\max_{e\in E'} f^i(\{e\})$, we have $\Delta_i( e(\pi^o,  \Psi^{o_{k-1}}, i) \mid \emptyset)= f^i(\{ e(\pi^o,  \Psi^{o_{k-1}})\}) \leq f^i(\{e(i)\})$. It follows that $\Delta_i( e(\pi^o,  \Psi^{o_{k-1}}, i) \mid \Psi^{o_{k-1}})\leq f^i(\{e(i)\})$ for all $\Psi^{o_{k-1}}$. Hence,
\begin{eqnarray}
\label{eq:tb3}\mathbb{E}_{\Psi^{o_{k-1}}, \Pi^o}[\Delta_i( e(\pi^o,  \Psi^{o_{k-1}}, i) \mid \Psi^{o_{k-1}})\mid \Psi^{o_{k-1}}] \leq f^i(\{e(i)\})
\end{eqnarray}
Now we are ready to prove this theorem.
\begin{eqnarray*}
&& m\times f_{avg}(\pi^{o}) = \sum_{i\in M} f^i_{avg}(\pi^{o})\\
&&=    \sum_{i\in M} f^i(S^o) +  \sum_{i\in M}\mathbb{E}_{\Psi^{o_{k-1}}, \Pi^o}[\Delta_i( e(\pi^o,  \Psi^{o_{k-1}}, i) \mid \Psi^{o_{k-1}})\mid \Psi^{o_{k-1}}]\\
&&\leq   e \sum_{i\in M} f^i(S^{a})+ \sum_{i\in M} f^i(\{e(i)\})\\
&&=  (1+e)(\frac{e}{1+e} \sum_{i\in M} f^i(S^{a})+ \frac{1}{1+e}\sum_{i\in M} f^i(\{e(i)\}))\\
&& = m\times  (1+e) f_{avg}(\pi^{a})
\end{eqnarray*}
The second equality is due to (\ref{eq:tb1}), the first inequality is due to (\ref{eq:tb3}), and the third equality is due to (\ref{eq:tb}).  Hence, $f_{avg}(\pi^{a}) \geq \frac{1}{1+e}f_{avg}(\pi^{o})$. $\Box$

\paragraph{A $\frac{1}{2e}$-approximate solution when $l=1$.} When $l=1$, we are allowed to select at most $l=1$ item at the training stage, and the remaining $k-1$ items can be selected adaptively in the test stage. We next propose a randomized policy $\pi^b$ that achieves a $\frac{1}{2e}$ approximation ratio to this case. $\pi^b$ does not select any items during the training set, i.e., the initial solution set chosen by $\pi^b$ is empty. After observing the incoming task, say $i\in G$, $\pi^b$ samples a policy uniformly at random from $\pi^{b1}$ and $\pi^{b2}$ to follow. We next describe the details of $\pi^{b1}$ and $\pi^{b2}$.
\begin{itemize}
\item The first candidate solution $\pi^{b1}$ selects $k-1$ items adaptively in a greedy manner: it starts with an empty set,  at each round $t\in[1, k-1]$ of $\pi^{b1}$,  it  selects an
item uniformly at random from the set $U(\psi^{b_{t-1}})$, where $U(\psi^{b_{t-1}}) = \arg\max_{V \in E': |V|\leq k-1}\sum_{e\in V} \Delta_i(e\mid  \psi^{b_{t-1}})$ contains the $k-1$ items with the largest marginal utility on top of the current partial realization $\psi^{b_{t-1}}$.
\item The second candidate solution $\pi^{b2}$ selects a singleton $e(i)$ with the largest expected utility.
\end{itemize}

We next analyze the performance bound of $\pi^{b}$.
\begin{theorem}
When $l=1$, $\pi^{b}$ achieves a $\frac{1}{2e}$ approximation ratio, i.e.,  $f_{avg}(\pi^{b}) \geq \frac{1}{2e}f_{avg}(\pi^{o})$.
\end{theorem}
\emph{Proof:}  Consider a one-step-further version $\pi^{b+}$  of $\pi^b$ by allowing it to select $k$ items in the test stage. Clearly, $f^i_{avg}(\pi^{b+}_{k-1})= f^i_{avg}(\pi^b)$ for any $i\in G$.  According to Theorem 1 in  \citep{tang2021beyond}, we can lower bound the performance of $\pi^{b+}$ as follows:
\begin{eqnarray}
\label{eq:super0}
f^i_{avg}(\pi^{b+})\geq \frac{1}{e} f^i_{avg}(\pi^{o})
  \end{eqnarray}
due to $f^i$ is adaptive submodular and $\pi^{o}$ is a feasible adaptive policy that selects at most $k$ items. Assume $e_k$ is the last item added to the solution by $\pi^{b+}$, we have $\Delta_i(e_k\mid \psi^{b_{k-1}})\leq \Delta_i(e_k\mid \emptyset)$ for any  $\psi^{b_{k-1}}$ due to $f^i$ is adaptive submodular and $\emptyset \subseteq \psi^{b_{k-1}}$. It follows that $\Delta_i(e_k\mid \psi^{b_{k-1}}) \leq \max_{e\in E'} \Delta_i(e\mid \emptyset)=f^i(\{e(i)\})$. Hence,
\begin{eqnarray}
\label{eq:super00}
f^i_{avg}(\pi^{b+}) \leq f^i_{avg}(\pi^{b+}_{k-1})+f^i(\{e(i)\}) = f^i_{avg}(\pi^{b})+f^i(\{e(i)\})
   \end{eqnarray}
(\ref{eq:super0}) and (\ref{eq:super00}) imply that
\begin{eqnarray}
\label{eq:super}
\frac{1}{e} f^i_{avg}(\pi^{o}) \leq f^i_{avg}(\pi^{b})+f^i(\{e(i)\})
 \end{eqnarray}
 Because $\pi^b$ samples a policy uniformly at random from $\pi^{b1}$ and $\pi^{b2}$ to follow, we have
 \begin{eqnarray}
\label{eq:super1}
f^i_{avg}(\pi^{b}) = (f^i_{avg}(\pi^{b})+f^i(\{e(i)\}))/2
 \end{eqnarray}
 (\ref{eq:super}) and (\ref{eq:super1}) together imply that $\frac{1}{e}  \sum_{i\in M} f^i_{avg}(\pi^{o}) = \sum_{i\in M}\frac{1}{e} f^i_{avg}(\pi^{o}) \leq \sum_{i\in M} (f^i_{avg}(\pi^{b})+ f^i(\{e(i)\}))=2 \sum_{i\in M} f^i_{avg}(\pi^{b})$. Hence, $f_{avg}(\pi^{b}) \geq \frac{1}{2e}f_{avg}(\pi^{o})$ due to $f_{avg}(\pi^{b})=\frac{1}{m}\sum_{i\in M} f^i_{avg}(\pi^{b})$ and
 $f_{avg}(\pi^{o})=\frac{1}{m}\sum_{i\in M} f^i_{avg}(\pi^{o})$.  $\Box$

\section{Performance Evaluation}

In this section, we evaluate the effectiveness of the proposed adaptive meta-learning strategy {\it Two-phase Greedy Policy} ({\it TGP}) and
compare with other benchmark approaches. Our experimental setup involves a set of tasks which are represented as submodular maximization problems subject to the $k$-cardinality constraint. We conduct experiments in the context of adaptive viral marketing. Given a social network represented by a directed graph, and a set of products, each task refers to promoting a particular product through a social network. As each product may have its own diffusion model that governs the diffusion process of this product, it is reasonable to select different sets of influential users (seeds) for marketing different products. We aim to select a set of seeds of size $k$ for each task to maximize the expected cascade in the social network over all products. As under the adaptive setting, we are allowed to choose the next seed after observing the actual spread resulting from previously selected seeds.

{\bf Dataset.} We conduct experiments on the benchmark dataset NetHEPT that is extensively used in many influence maximization studies \citep{chen2016robust,sun2018multi}. It is an academic collaboration network extracted from the High Energy Physics Theory section of arXiv from 1991 to 2003. The nodes represent the authors and each edge represents the collaboration of two authors on a paper. The graph contains $15,233$ nodes and $62,774$ directed edges. The propagation probability of each directed edge is sampled randomly from $\{0.1, 0.01\}$ as in \citep{yuan2017no}. For training we form $m=50$ tasks by generating for each task an assignment of propagation probability of all edges in the graph. We test on $m=50$ new tasks sampled from the same distribution and report in the figures the average performance obtained on test tasks.

\begin{figure*}[tb]
\hspace*{-0.9in}
\centering
\includegraphics[scale=0.21]{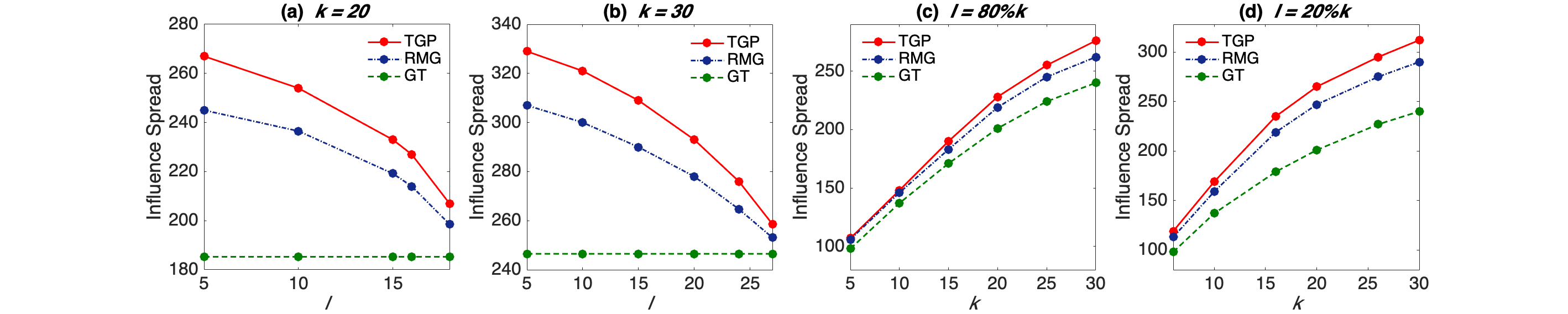}
\caption{Performance for Adaptive Influence Maximization under different settings of $l$ and $k$.}
\label{fig:result}
\end{figure*}

{\bf Algorithms.} We evaluate the performance of {\it TGP} compared with two baselines. {\it Greedy Train} ({\it GT}) chooses all the $k$ items greedily during the training phase. {\it GT} is a special case of {\it TGP} where $l=k$, i.e., this case has zero degree of personalization. {\it Randomized Meta-Greedy} ({\it RMG})  is the state-of-the-art non-adaptive submodular meta-learning algorithm developed in  \citep{adibi2020submodular}. Following the framework of meta-learning, {\it RMG} first finds an initial solution of size $l$ at training time and the solution is then completed for each test task \emph{non-adaptively}, i.e., one is not allowed to observe the partial realization during the selection process. All experiments were run on a machine with Intel Xeon 2.40GHz CPU and 16GB memory, running 64-bit RedHat Linux server. For each set of experiments, we run the simulation for 100 rounds and average results are reported as follow.

{\bf Results.} Figure \ref{fig:result} shows the performance of our proposed algorithm against the baselines. Figure \ref{fig:result}(a) shows the performance of the algorithms when we fix $k=20$, and vary $l$ from $5$ to $18$. Larger $l$ means less computation at test time as we need to add $k-l$ seeds at test. We observe that as $l$ decreases, the influence spread produced by {\it TGP} and {\it RMG} both increase. It indicates that adding a few personalized items at test time significantly boosts performance. We also observe that {\it TGP} outperforms {\it RMG} in terms of influence spread, since the latter does not utilize the observation that may be made during the seeding process at test time. Figure \ref{fig:result}(b) shows the performance of the algorithms when we fix $k=30$, and vary $l$ from $5$ to $27$. We observe a very similar pattern as in Figure \ref{fig:result}(a). In Figure \ref{fig:result}(c), we compare the performance of the algorithms when $k$ changes from $5$ to $30$, and $l$ is $80\%$ of $k$ ($l=0.8k$). As we can see, the influence spread increases for all algorithms as $k$ increases. As expected, {\it TGP} outperforms {\it RMG} and {\it GT} on all test cases. And the performance gap between {\it TGP} and {\it RMG} also increases with $k$. Figure \ref{fig:result}(d) has been obtained in a similar format as Figure \ref{fig:result}(c) only with $l$ set to $20\%$ of $k$ ($l=0.2k$) instead. Again, {\it TGP} outperforms {\it RMG} and {\it GT} on all test cases, and the performance gap between {\it TGP} and {\it RMG} increases as $k$ increases.

\section{Conclusion}
In this paper, we develop a novel framework of adaptive submodular meta-learning. We extend the notion of submodular meta-learning to the adaptive setting which allows each item to have a random state.  Our goal is to find an initial set of items that can quickly adapt to a new task. We propose a two-phase greedy policy that achieves a $1/2$ approximation ratio for the monotone case. For the non-monotone case, we proposed a two-phase randomized greedy policy that achieves a $1/32$ approximation ratio. We evaluated the performance of our proposed algorithm
for the application of adaptive viral marketing.

\bibliographystyle{ijocv081}
\bibliography{reference}




\end{document}